\documentclass[runningheads]{llncs}
\usepackage{graphicx}
\usepackage{amsmath,amssymb} 
\usepackage{color}
\usepackage{booktabs}
\usepackage{subfig}
\usepackage{url}

\begin{document}

\title{AttentionMask: Attentive, Efficient Object Proposal Generation Focusing on Small Objects} 
\titlerunning{AttentionMask} 


\author{Christian Wilms \and Simone Frintrop}
%

\authorrunning{C. Wilms and S. Frintrop} 


\institute{University of Hamburg\\
\email{\{wilms,frintrop\}@informatik.uni-hamburg.de}}

\maketitle

\begin{abstract}
We propose a novel approach for class-agnostic object proposal generation, which is efficient and especially well-suited to detect small objects. Efficiency is achieved by scale-specific objectness attention maps which focus the processing on promising parts of the image and reduce the amount of sampled windows strongly. This leads to a system, which is $33\%$ faster than the state-of-the-art and clearly outperforming state-of-the-art in terms of average recall. Secondly, we add a module for detecting small objects, which are often missed by recent models. We show that this module improves the average recall for small objects by about $53\%$. Our implementation is available at: \url{https://www.inf.uni-hamburg.de/en/inst/ab/cv/people/wilms/attentionmask}.
\end{abstract}

\section{Introduction}
\label{sec:intro}
Generating class-agnostic object proposals is an important part of many state-of-the-art object detectors \cite{girshick2014rich,he2014spatial,ren2015faster,girshick2015fast,redmon2016you} as well as instance segmentation systems \cite{li2016fully,he2017mask,chen2018masklab}, since it reduces the number of object candidates to a few hundreds or thousands. The results of object proposal systems are either bounding boxes or pixel-precise segmentation masks. Both streams have received a massive boost recently due to the rise of deep learning \cite{ren2015faster,Pinheiro2015-deepmask,Pinheiro2016-sharpmask,qiao2017scalenet,Hu2017-fastmask,li2017zoom}. The task of generating class-agnostic object proposals is different from object detection or instance segmentation as those systems predict a class label and use this information when generating boxes or masks \cite{ren2015faster,li2016fully,he2017mask,chen2018masklab}. Thus, they generate class-specific and not class-agnostic results.

\begin{figure}[b]
    \centering
    \subfloat{\includegraphics[width = .5\linewidth]{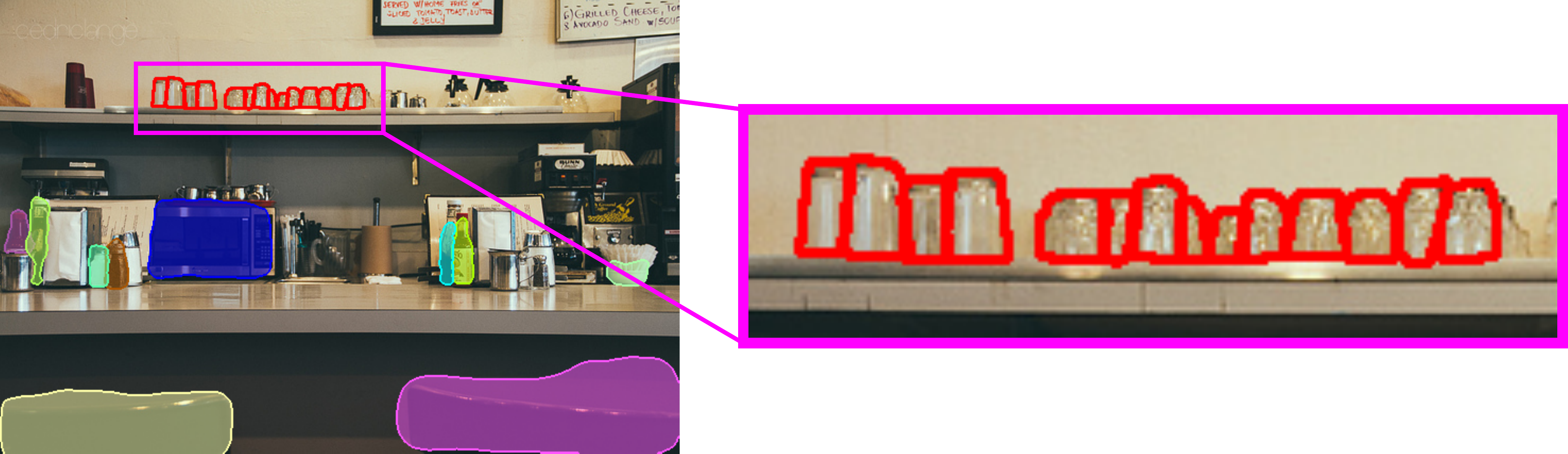}}~
    \subfloat{\includegraphics[width = .5\linewidth]{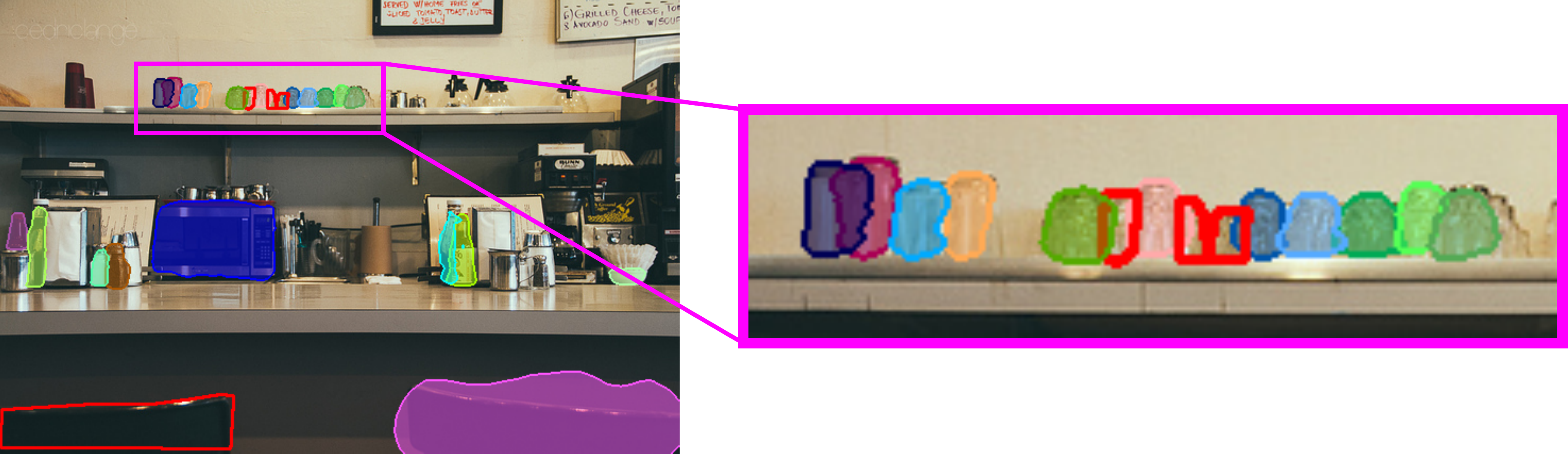}}
            
    \caption{Result of FastMask~\cite{Hu2017-fastmask} (left) and the proposed AttentionMask (right) on small objects, highlighting one strength of AttentionMask. The filled colored contours denote found objects, while the red contours denote missed objects.}\label{fig:qualResCrop1}
\end{figure}

Despite the impressive performance gain since the arrival of deep learning based class-agnostic object proposal systems, some major problems still persist. For example, most systems perform well on medium or large sized objects, but exhibit a dramatic drop in performance on small objects \cite{Pinheiro2015-deepmask,Pinheiro2016-sharpmask,Hu2017-fastmask}. See Fig.~\ref{fig:qualResCrop1} for an example. However, in our daily lives we are surrounded by many small objects, e.g., pens, a computer mouse, or coins. The significant loss when trying to detect small objects originates mainly from subsampling the original image in order to reduce the computational complexity.

A second problem of most recent class-agnostic object proposal systems is an inefficient use of the resources. Most systems use an image pyramid and sample many windows that have to be processed individually \cite{Pinheiro2015-deepmask,Pinheiro2016-sharpmask,qiao2017scalenet}. Although \cite{qiao2017scalenet} optimize the scale selection, there is a lot of redundancy with overlapping windows and multiple scales. Hu et al.~\cite{Hu2017-fastmask} took a first step by transferring the sampling of windows to the feature space and thus feeding the input image only once through the base net. However, still many unnecessary windows are sampled in background regions that can be easily omitted. Humans, for instance, have no problem to quickly ignore background regions as they developed the mechanism of visual attention to focus processing on promising regions \cite{Pashler97}. 

\begin{figure}[tbp]
	\centering
      \includegraphics[width=1\linewidth]{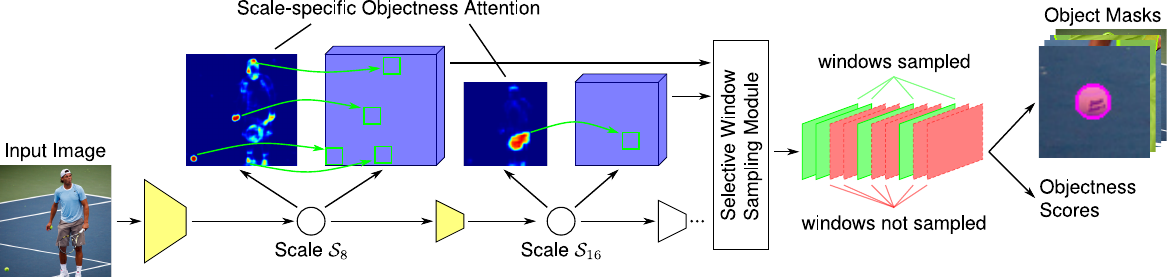}
	  \caption{Overview of the proposed AttentionMask system: An image is fed through the base net (yellow) and further downscaled to different scales ($\mathcal{S}_8,\mathcal{S}_{16},\dots$). At each scale a scale-specific objectness attention is calculated and windows (green) are sampled in the feature maps (blue boxes) at the most promising locations only to make calculations more efficient. Windows pruned due to low attention are marked in red. Each window is scored and an object mask is created. Due to the more efficient usage of the resources by introducing scale-specific objectness attention, we are able to split the base net and add a module ($\mathcal{S}_8$) to better detect small objects like the tennis balls in this example.}
	\label{fig:abstractOverview}
	\end{figure}

In this paper, we propose AttentionMask, a new deep learning system to generate class-agnostic object proposals in a more efficient way, which opens the chance to process more scales and thus enables us to better detect smaller objects. AttentionMask is based on the FastMask architecture \cite{Hu2017-fastmask}, which is extended in two essential ways: first, to increase the efficiency, we use the bio\-logi\-cally-inspired approach of visual attention~\cite{Pashler97} and include \emph{scale-specific objectness attention maps}. This attention focuses the sampling of windows on promising parts of the image, i.e., containing objects of a specific scale, (green windows in Fig.~\ref{fig:abstractOverview}) while omitting sampling of windows in background regions (red windows in Fig.~\ref{fig:abstractOverview}). Second, taking advantage of the lower consumption of resources in terms of both memory and processing time, we are able to add an additional module ($\mathcal{S}_8$ in Fig.~\ref{fig:abstractOverview}). This new module extracts features earlier in the network, which suffer less from the subsampling (cf.~Fig.~\ref{fig:abstractOverview}). Our results show that the scale-specific objectness attention significantly increases the efficiency in runtime and memory consumption as well as improving results considerably. Furthermore, adding this module to the system drastically increases the chance of detecting small objects as visible in Fig.~\ref{fig:qualResCrop1}.

Our major contributions are as follows:
\begin{enumerate}
\item Introducing a scale-specific objectness attention to omit background parts of the image and increase efficiency ($33\%$ faster).
\item Adding a new module to the system for detecting small objects by taking advantage of the more efficient processing.
\item Introducing an efficient system that clearly outperforms state-of-the-art-methods ($+8\%$ to $+11\%$), especially on detecting small objects ($+53\%$).
\end{enumerate}

The rest of the paper is organized as follows. Sec.~\ref{sec:relatedWork} gives an overview over related work in class-agnostic object proposal generation and visual attention. In Sec.~\ref{sec:fastMask}, we introduce the baseline network and in Sec.~\ref{sec:methods} our key methodological contributions. Sec.~\ref{sec:impl} gives an overview over technical details of our approach, while in Sec.~\ref{sec:results} results and an evaluation against state-of-the-art methods are presented. The paper closes with concluding remarks in Sec.~\ref{sec:conclusion}.

\section{Related Work}
\label{sec:relatedWork}
This section gives an overview over the most relevant work in the field of class-agnostic object proposal generation and visual attention. 

\subsubsection{Object Proposal Generation.}
\label{sec:rWObjectProposals}
Since the seminal work of Alexe et al.~\cite{Alexe12PAMI} on class-agnostic object proposal generation, many systems, both hand-crafted \cite{HMFL15ICRA,Werner2015ICVS,pont2017multiscale,uijlings2013selective} and deep learning based \cite{Pinheiro2015-deepmask,Pinheiro2016-sharpmask,qiao2017scalenet,Hu2017-fastmask}, have been proposed to generate bounding boxes or pixel-precise segmentation masks of the object proposals. Hosang et al.~\cite{Hosang2015PAMI} give a good overview over early hand-crafted work.

Bounding box proposal systems score many sliding windows to select the most promising ones using hand-crafted features  \cite{Alexe12PAMI,zitnick2014edgeboxes,cheng2014bing} or deep learning~\cite{ren2015faster,li2017zoom}. Closest to the presented system are \cite{zitnick2014edgeboxes}, \cite{cheng2014bing}, and \cite{li2017zoom}. \cite{zitnick2014edgeboxes} and \cite{cheng2014bing} use hand-crafted features to generate proposals with low runtime. \cite{li2017zoom} combine features from early and final layers of a network to better detect small objects. In contrast to those works, we generate pixel-precise proposals, which is more challenging.

A slightly different objective than regressing boxes is the generation of pixel-precise segmentation masks. The systems either group segments together using a variety of strategies to from a diverse set of proposals~\cite{uijlings2013selective,Werner2015ICVS} or use CNN-based methods~\cite{Pinheiro2015-deepmask,Pinheiro2016-sharpmask,qiao2017scalenet,Hu2017-fastmask,dai2016instance}, which achieve better results. DeepMask \cite{Pinheiro2015-deepmask} and SharpMask \cite{Pinheiro2016-sharpmask} rely on a patch-based processing of the image. For each patch, a segmentation mask as well as an objectness score is computed. SharpMask, in contrast to DeepMask, uses refinement modules for more precise masks. Still, an image has to be fed through the system in multiple scales and multiple patches.  To estimate the scales that are relevant for the content, \cite{qiao2017scalenet} developed ScaleNet. To not feed the same image parts multiple times through the base net, Hu~et~al.~\cite{Hu2017-fastmask} propose FastMask. FastMask feeds the image only once through the base net at one scale and derives a feature pyramid. On the resulting feature maps of different scales, windows are sampled to find objects of different sizes. FastMask, explained in more detail in Sec.~\ref{sec:fastMask}, serves as the base for our AttentionMask. AttentionMask differs from the aforementioned using a notion of visual attention to sample windows only in the most promising locations of the feature maps and as a result freeing up resource to be able to focus more on finding small objects.

\subsubsection{Visual Attention.} 
Visual attention is a mechanism of human perception that focuses the processing in the brain on the most promising parts of the visual data~\cite{Pashler97}. The concept of visual attention has been studied in psychology and neuroscience for decades and early computational models of visual attention~\cite{tsotsos93,itti_koch_niebur98} have been inspired by psychological models~\cite{treisman80,wolfe89}. Later, many different computational models of visual attention have been proposed \cite{harel07,Frintrop15CVPR,kleinFrintrop_iccv2011,huang2015salicon,hou2017deeply}. 

In addition to these approaches which explicitly model visual attention, a recent trend uses the term ``attention'' in a more general sense, indicating that the processing in a deep network is concentrated on specific image regions for very different tasks such as visual question answering \cite{yang2016stacked} or image captioning \cite{xu2015show,you2016image,chen2017sca}. Most related to our approach are the works of \cite{chen2016attention} and \cite{kong2017ron}. \cite{chen2016attention} learn to weight features of different scales for semantic segmentation computing attention for each scale. \cite{kong2017ron} use attention of different scales to determine promising bounding boxes for object detection. Both approaches address different problems than we do: \cite{chen2016attention} addresses semantic segmentation while \cite{kong2017ron} works on object detection. 

\section{Baseline Network: FastMask}
\label{sec:fastMask}
Since the core structure of AttentionMask is based on FastMask~\cite{Hu2017-fastmask}, we briefly explain the model here for completeness. FastMask is a CNN-based system that generates pixel-precise class-agnostic proposal masks. The main idea behind FastMask is to increase the efficiency by feeding the image only once through the base net and do both, constructing a scale pyramid and sampling windows in feature space. This approach is called one-shot paradigm. In contrast, most CNN-based proposal systems follow the multi-shot paradigm, constructing an image pyramid and sample windows in image space, which are individually passed through the network \cite{Pinheiro2015-deepmask,Pinheiro2016-sharpmask,qiao2017scalenet}. This leads to redundancy as the windows partially overlap or contain the same image area in different scales. The one-shot paradigm removes this redundancy in the base net to speed up the process. 

Following this one-shot paradigm, FastMask uses a ResNet~\cite{he2016-resnet} as base net with four strided convolutions (factor 2) to compute an initial feature map of the input image that is downscaled by a factor of $16$ compared to the input. This feature map is further downscaled multiple times by the factor $2$ using \textit{residual neck modules} introduced by \cite{Hu2017-fastmask} yielding a feature pyramid. From those feature maps, all possible windows of a fixed size are sampled. Using the same fixed size on feature maps of different scales leads to windows of different size in the original image, i.e., objects of different size can be found. Working with 4 scales, i.e., downscaling the image through the base net and having three neck modules, results in the scales $16$, $32$, $64$ and $128$ denoted as $\mathcal{S}_{16},\mathcal{S}_{32},\mathcal{S}_{64}$, and $\mathcal{S}_{128}$ respectively. Thus, each window is a feature representation of a patch of the image with four different patch sizes. To be more robust to object sizes, FastMask has a second optional stream with $\mathcal{S}_{24},\mathcal{S}_{48},\mathcal{S}_{96}$, and $\mathcal{S}_{192}$, obtained by a new branch using stride $3$ in the last strided convolution of the base net.

The windows of the different scales form a batch of feature maps. Each of the windows is fed into a sub-network that calculates the likelihood of an object being centered and fully contained in the window. The most promising $k$ windows are further processed to derive a segmentation mask of that object. Since the aspect ratio of the windows is always the same and the objects are not necessarily centered, an \textit{attentional head} is introduced to prune features in irrelevant parts of the windows before segmentation. The final result are $k$ pixel-precise masks with according objectness scores.

\section{Methods} 
\label{sec:methods}
\begin{figure}[tbp]
	\centering
      \includegraphics[width=1\linewidth]{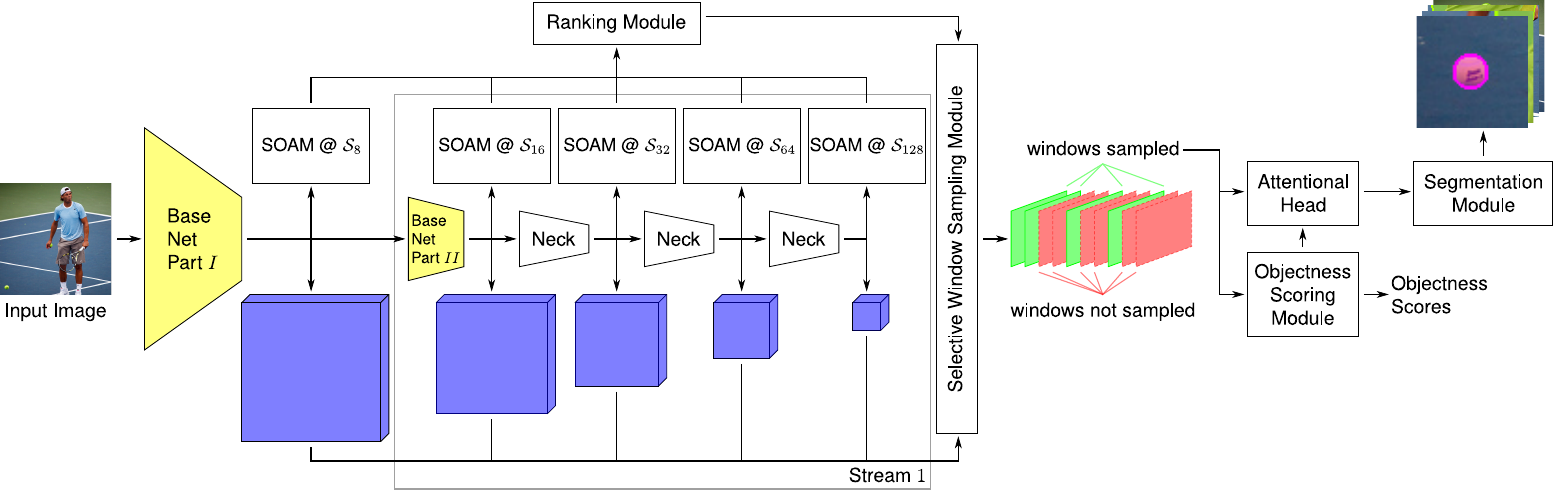}
	  \caption{Overview of the proposed AttentionMask. The base net (yellow) is split to allow $\mathcal{S}_8$. At each scale ($\mathcal{S}_n$), features are extracted (blue boxes) and a scale-specific objectness attention module (SOAM) computes an attention map. Based on the joint ranking of the attention values, windows are only sampled in the most promising areas. Those windows are fed through the objectness scoring module, to generate objectness scores as well as the attentional head and the segmentation module to generate pixel-precise class-agnostic segmentations.}
	\label{fig:overview}
	\end{figure}

In this section, we introduce AttentionMask, visualized in Fig.~\ref{fig:overview}, a novel deep learning system generating pixel-precise class-agnostic object proposals. While the core structure is based on FastMask (cf. Sec.~\ref{sec:fastMask}), our approach is essentially different in two aspects: first, it adds attentional modules at every scale that generate scale-specific objectness attention maps and enables a strong reduction of the sampled windows and thus a more efficient processing. Second, we add a module to the early stage of the base net to improve the detection of smaller objects. Sampling more windows from the new module is only feasible, since we reduce the number of windows sampled per scale using the scale-specific objectness attention maps. The following sections first describe the general architecture of our system and then focus on the two major novel components in more detail.

\subsection{General Overview}
\label{sec:generalOverview}
Similarly to FastMask and in contrast to DeepMask \cite{Pinheiro2015-deepmask} and SharpMask \cite{Pinheiro2016-sharpmask}, AttentionMask follows the one-shot paradigm, thus feeding the image only once through the base net. The base net is a ResNet, which we split into two parts before the last strided convolution. This leads to an additional scale $\mathcal{S}_8$ that we use for better detection of small objects (see Sec.~\ref{sec:smallObjects}). The resulting feature map of the second part of the base net is further downscaled from $\mathcal{S}_{16}$ to $\mathcal{S}_{32},\mathcal{S}_{64}$, and $\mathcal{S}_{128}$ using residual neck components~\cite{Hu2017-fastmask}. Before sampling the windows, we introduce the new scale-specific objectness attention modules at each scale, which compute an attention value for each position of the feature maps, determining whether the location is part of an object of the scale (Sec.~\ref{sec:scaleSpecAtt}). According to the joint ranking of those attention values across the scales, windows are sampled only at the most promising locations in the feature maps with a new selective window sampling module. The resulting batch of windows is fed into an objectness scoring sub-network \cite{Hu2017-fastmask} to generate objectness scores. Based on this score, the most promising $k$ windows are processed by the attentional head \cite{Hu2017-fastmask} to prune background features and a segmentation module. Finally, all segmentation masks are scaled back to the size of the according windows in the original image to get pixel-precise class-agnostic segmentation masks.

\subsection{Scale-specific Objectness Attention}
\label{sec:scaleSpecAtt}
The first major novel component of AttentionMask that we propose is a scale-specific objectness attention module (SOAM) to focus the sampling of sliding windows at the most promising locations. These are regions that are likeliest to fully contain an object. In contrast to salient object detection systems, where attention is used to focus on the most salient object, here we use a specialized scale-specific objectness attention to focus on all objects of a certain scale. Different from the attentional head of FastMask, this module enables the system to significantly reduce the amount of sampled windows by focusing computation on the promising windows early in the network, which makes processing more efficient. As visible in Fig.~\ref{fig:overview}, the SOAMs are integrated between the features of the different scales and the ranking module. Thus, the SOAMs take the output of the base net or a further downscaled version and produce a scale-specific objectness attention map for each scale. Since now an attention value is assigned to every location in the feature maps, the locations across all scales can be ranked jointly according to their attention in the ranking module to find the overall most promising locations for sampling windows. This ranking is used by the subsequent selective window sampling module to sample windows only at the most promising locations that likely contain an object. 

\begin{figure}[tbp]
	\centering
      \includegraphics[width=1\linewidth]{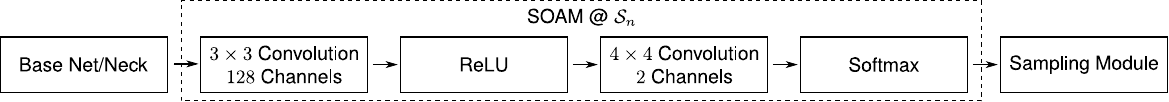}
	  \caption{Detailed view of our scale-specific objectness attention module (SOAM).}
	\label{fig:soam}
	\end{figure}

\begin{figure}[tb]
\centering
\begin{tabular}{cccccc}
\subfloat[Input]{\includegraphics[width = .16\linewidth]{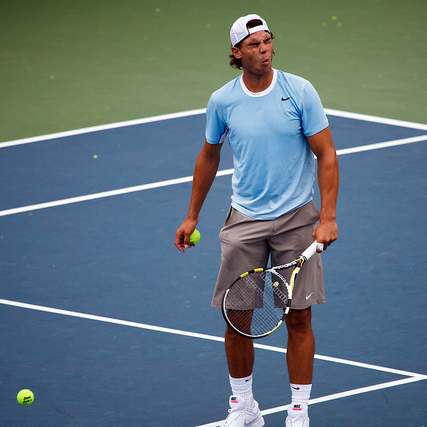}} &
\subfloat[$\mathcal{S}_{8}$]{\includegraphics[width = .16\linewidth]{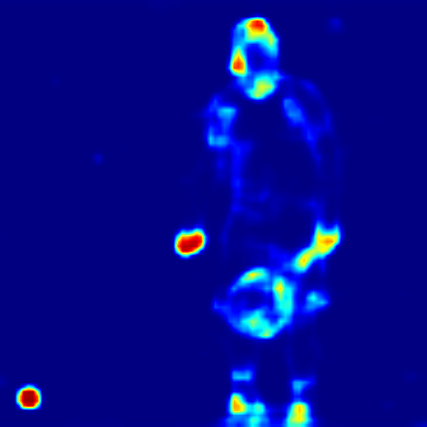}} &
\subfloat[$\mathcal{S}_{16}$]{\includegraphics[width = .16\linewidth]{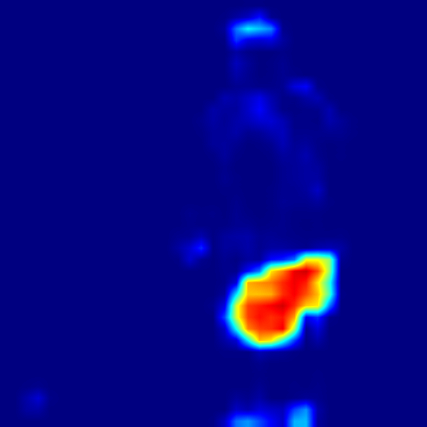}}&
\subfloat[$\mathcal{S}_{32}$]{\includegraphics[width = .16\linewidth]{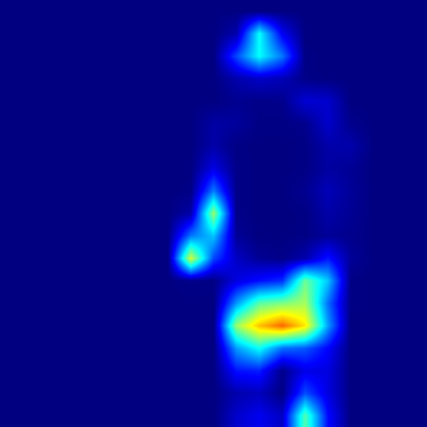}} &
\subfloat[$\mathcal{S}_{64}$]{\includegraphics[width = .16\linewidth]{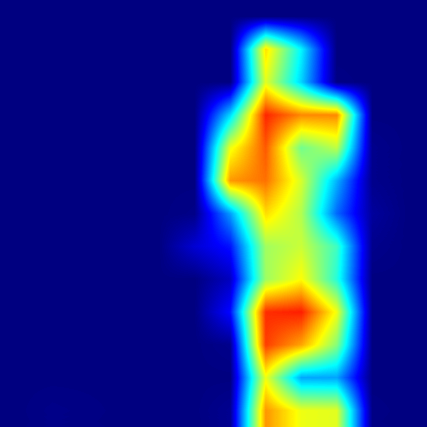}} &
\subfloat[$\mathcal{S}_{128}$]{\includegraphics[width = .16\linewidth]{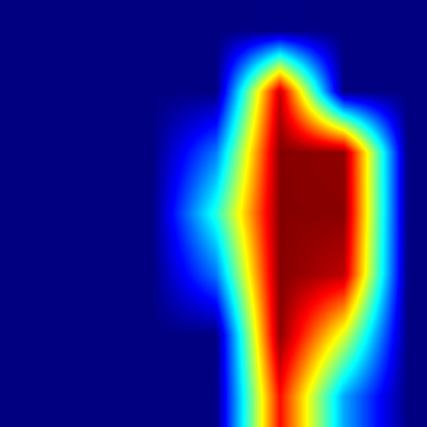}}
\end{tabular}
\caption{Scale-specific objectness attention maps for an image. The different scales highlight objects or object parts of different size. Only due to the new scale $\mathcal{S}_{8}$ we are able to detect the two tennis balls. The maps are rescaled to image size for better visibility.
}
\label{fig:attExample}
\end{figure}

We view the problem of calculating the scale-specific objectness attention maps as a semantic segmentation problem with only two classes, \emph{object} and \emph{non-object}. For the design of the SOAMs we follow the approaches of \cite{chen2017deeplab,long2015fully} on semantic segmentation using a fully-convolutional solution. Our approach, visualized in Fig.~\ref{fig:soam}, is to calculate the scale-specific objectness attention from the features of a scale using a $3 \times 3$ convolutional layer with $128$ channels and ReLU activation and a $4 \times 4$ convolutional layer with $2$ channels, i.e., one for each of the classes (the size of $4\times 4$ is needed as the windows are sampled with an evenly sized $10\times 10$ mask). This configuration has the best trade-off between efficiency and effectiveness, as the results in Tab.~\ref{tab:ablationAtt} indicate. Finally, softmax is applied to get the probabilities and as a result the scale-specific objectness attention map, which is the probability of the class \emph{object}. As a result of the SOAMs, we are now able to prioritize the potential windows as well as only sample and process the most promising candidates, which saves time and memory as all sampled windows form one batch for the subsequent computation. Fig.~\ref{fig:attExample} shows an example of the calculated scale-specific objectness attention maps. For details about the training procedure, see Sec.~\ref{sec:training}.

\textbf{How are the SOAMs different to the attentional head of FastMask?} The attentional head selects within a sampled window the most relevant location for the central object, while our SOAMs select the most promising locations to sample windows in the feature maps of different scales. The aim of the attentional head is to remove background clutter that might prevent a proper segmentation of the object, while the SOAMs aim to reduce the number of sampled windows for increased efficiency. Therefore, the modules operate in very different stages of the network and follow different objectives.

\textbf{Why not compute one attention map for all scales?} Calculating one objectness attention map for all scales, though, would lead to many windows sampled on large scales like $\mathcal{S}_8$, which only show parts of larger objects. Therefore, using a distinct objectness attention map for each scale is key to remove obvious false positives and reduce the number of windows. For results using only one objectness attention map see Tab.~\ref{tab:ablationAtt} row 2. 

\begin{table}[tb]
\parbox{.48\linewidth}{
\centering
\caption{Approaches for calculating scale-specific objectness attention.}
\label{tab:ablationAtt}
\begin{tabular}{@{}p{0.65\linewidth}cc@{}}
\toprule
 & AR@100 & Time \\ \midrule
$4\times4, 2 \text{ channels}$ &  0.258  & $0.20s$ \\
$4\times4, 2 \text{ ch.}$, only one scale &  0.216  & $0.25s$ \\ 
\textbf{$\mathbf{3\times3, 128 }\text{ ch.}\mathbf{ + 4\times4, 2} \text{ ch.}$}  &  \textbf{0.261}  & $\mathbf{0.21s}$ \\
$3\times3, 256 \text{ ch.} + 4\times4, 2 \text{ ch.}$ &  0.261  & $0.23s$ \\ \bottomrule
\end{tabular}
}
\hfill
\parbox{.48\linewidth}{
\centering
\caption{Approaches for adding $\mathcal{S}_8$ to AttentionMask.}
\label{tab:ablationS8}
\begin{tabular}{@{}p{0.6\linewidth}cc@{}}
\toprule
 & AR@100 & Time \\ \midrule
dilated convolution (i) & 0.292 & $0.30s$ \\
\textbf{direct connection (ii)} & \textbf{0.287} & $\mathbf{0.21s}$ \\
reverse connection (iii) & 0.285 & $0.26s$ \\ \bottomrule
\end{tabular}
}
\end{table}

\subsection{Generating Proposals for Small Objects}
\label{sec:smallObjects}
As a second novelty of AttentionMask, we introduce a module for an additional larger scale ($\mathcal{S}_8$) to improve detection of small objects. Small objects are regularly missed when only using $\mathcal{S}_{16}$ and subsequent scales as small objects do not cover enough area inside the sampled windows. Adding $\mathcal{S}_8$ however, leads to a large amount of new windows. For an input image of size $800 \times 600$, about 7500 windows are additionally sampled, more than in all subsequent scales together. To keep the efficiency of the system, scale-specific objectness attention is essential to reduce the amount of sampled windows  and keep the problem feasible. Therefore, we propose to use the scale-specific objectness attention and utilize the freed resources to add another scale to the top of the feature scale space, i.e., $\mathcal{S}_8$, to foster detection of small objects. Without the scale-specific objectness attention this would not be possible, due to memory limitations on modern GPUs.

Since the output of the base net of FastMask is a feature map of $\mathcal{S}_{16}$, there is no natural $\mathcal{S}_8$ yet. We investigated three ways of calculating a $\mathcal{S}_8$ feature map in AttentionMask, inspired by different approaches. (i) One way of preventing the resolution of the feature maps of a CNN from decreasing with increasing depth is the use of dilated convolution as in \cite{chen2017deeplab,yu2015multi}. Inspired by the success of those systems in semantic segmentation, we replaced the last strided convolution by a dilated convolution with factor $2$ in this and all subsequent layers to keep the resulting feature map at $\mathcal{S}_8$. Despite a good performance, see Tab.~\ref{tab:ablationS8}, the dilated convolutions slow down the network significantly ($+36\%$). (ii) Inspired by \cite{kong2017ron}, we directly extract features before the last strided convolution. Despite not having the full depth of the base net, those features are still relevant for small objects, as those are usually less complex due to their small size in the image. (iii) We also evaluated the reverse connection approach presented in \cite{kong2017ron}. The reverse connection here is a deconvolutional layer connecting $\mathcal{S}_{16}$ with $\mathcal{S}_8$. Through this reverse connection, information from $\mathcal{S}_{16}$ can flow back to $\mathcal{S}_8$.

As shown in Tab.~\ref{tab:ablationS8}, the second approach of directly extracting features before the last strided convolution performs well, while leading to the fastest execution time. Therefore, we will use this approach in the following.

\textbf{Why not directly add $\mathcal{S}_8$ to FastMask?} One of the problems of FastMask is the memory consumption, especially in training. Adding $\mathcal{S}_{192}$, e.g., for the above mentioned input image of size $800 \times 600$ results in only 20 more windows. As described above, adding $\mathcal{S}_8$ leads to around 7500 additional windows. Memory usage would prevent the system from being trained on modern GPUs. Using a further neck is also not possible as they can only downscale but not upscale.

\section{Implementation Details}
\label{sec:impl}
In this section, we present technical details of AttentionMask splitting the base net, connecting the SOAMs with the new selective window sampling module and variations of AttentionMask. Furthermore, details about training are given.

\subsection{Architecture Details} 
\label{sec:arch}
Adding $\mathcal{S}_{8}$ to better detect small objects (Sec.~\ref{sec:smallObjects}) leads to splitting up the base net. We use as our base net a ResNet-50 without the \textit{conv5}-module. Given the inherent structure of the stripped ResNet, we split the network between the \textit{conv3-} and \textit{conv4}-modules, thus moving the final strided convolution to the second part of the base net. This enables us to extract features of the \textit{conv3d}-layer, the final layer of the \textit{conv3}-module, for $\mathcal{S}_{8}$. 

To use the scale-specific objectness attention introduced in Sec.~\ref{sec:scaleSpecAtt} for selecting windows, all attention maps have to be ranked jointly. This is done by the new ranking module that jointly ranks the locations in the maps across the different scales. This ranking is used in the new selective window sampling module to only sample windows at the most promising locations in the feature maps.

Similar to \cite{Hu2017-fastmask}, we add a second stream to AttentionMask, increasing the number of scales from 5 to 9. This is done by duplicating the \textit{conv4}-module of the ResNet and use strided convolutions with stride 2 and 3, which leads to scales $\mathcal{S}_{16}$ (stream 1) and $\mathcal{S}_{24}$ (stream 2). However, \cite{Hu2017-fastmask} are not able to train this model entirely and have to transfer weights between the necks. In contrast to that, we are also able to train the full model with two streams and 9 scales, due to our efficient attention modules. In total, we run 3 versions of the model distinguished by the scales used: (i) AttentionMask$^{16}_{192}$ uses  the scales $\mathcal{S}_{16}$ to $\mathcal{S}_{192}$, which corresponds to the model of \cite{Hu2017-fastmask} just adding attention. (ii) AttentionMask$^{8}_{192}$ adds $\mathcal{S}_{8}$ to the previous model resulting in 9 scales. (iii) To compare AttentionMask with other models having eight scales \cite{Pinheiro2015-deepmask,Pinheiro2016-sharpmask,Hu2017-fastmask} we use a third version with $\mathcal{S}_{8}$ to $\mathcal{S}_{128}$ (AttentionMask$^{8}_{128}$). Furthermore, \cite{Hu2017-fastmask} proposed to use PVANet~\cite{hong2016pvanet} as base net to make inference faster. However, as the model from \cite{Hu2017-fastmask} is not publicly available, we did not include a version using PVANet in the paper.

\subsection{Joint Training}
\label{sec:training}

Training AttentionMask consists of multiple different tasks and thus multiple losses. Training the baseline network follows \cite{Hu2017-fastmask} and uses three losses for the objectness score $\mathcal{L}_{objn}$, the attentional head output $\mathcal{L}_{ah}$ and the segmentation mask $\mathcal{L}_{seg}$ respectively. See the supplementary material for further details. For training the SOAMs, we calculate the softmax over the output and compute the loss $\mathcal{L}_{att_n}$ at $\mathcal{S}_n$. The ground truth is created based on the ground truth masks for all objects that fit to the scale $\mathcal{S}_n$ of a scale-specific objectness attention map. An object fits to $\mathcal{S}_n$, if both side lengths of the object are within $40\%$ to $80\%$ of the sampled window side length of $\mathcal{S}_n$ in the original image. For handling the significant imbalance in the ground truth at different scales, we follow \cite{kong2017ron} by sampling for each positive pixel in a ground truth map 3 negative pixels. A comparison with other strategies is presented in the supplementary material. 

Thus, the overall loss is
\begin{equation}
\mathcal{L} = w_{objn}\mathcal{L}_{objn}+w_{ah}\mathcal{L}_{ah}+w_{seg}\mathcal{L}_{seg} + w_{att} \sum_n \mathcal{L}_{att_n},
\end{equation}
with $w_{objn}, w_{ah}, w_{seg}$, and $w_{att}$ being weights to balance the influence of the tasks. In our experiments we set $w_{objn}=0.5, w_{ah}=1.25, w_{seg}=1.25$ and, $w_{att}=0.25$, as it gave the best results. Note that during training, there is no connection between the SOAMs and the selective window sampling module. Instead, the ground truth data is used to sample windows, as it leads to a better performance (evaluation in supplementary material). Details regarding the hyperparameters and solver can be found in the supplementary material.

\section{Experiments}
\label{sec:results}
This section presents results and the evaluation of AttentionMask compared to several state-of-the-art methods in terms of performance and efficiency as well as ablation experiments analyzing AttentionMask in more detail. The evaluation is carried out on the MS COCO dataset \cite{lin2014microsoft}. The MS COCO dataset has more than 80.000 training images with pixel-precise annotations of objects. As done by \cite{Pinheiro2015-deepmask,Pinheiro2016-sharpmask,Hu2017-fastmask,dai2016instance}, we use the first 5.000 images of the validation set for testing.  Training was conducted on the training set of MS COCO. 

Following \cite{Hosang2015PAMI,Pinheiro2015-deepmask,Pinheiro2016-sharpmask,Hu2017-fastmask,dai2016instance}, we report average recall (AR) at 10, 100 and 1000 proposals, which correlates well with the performance of object detectors using these proposals~\cite{Hosang2015PAMI}. Additionally, we separate our evaluation with respect to the performance for different object sizes. We follow the MS COCO classification into small objects (area $<32^2$), medium objects ($32^2<$ area $<96^2$), and large objects ($96^2<$ area) and report the AR with respect to those classes at 100 proposals. We further report the inference runtime when generating 1000 proposals. All experiments, except for InstanceFCN (numbers taken from \cite{Hu2017-fastmask}), were carried out in a controlled environment with a NVIDIA GeForce GTX TITAN X.

The state-of-the-art methods we compare to are FastMask \cite{Hu2017-fastmask}, DeepMask\-Zoom \cite{Pinheiro2015-deepmask,Pinheiro2016-sharpmask}, SharpMask as well as SharpMaskZoom \cite{Pinheiro2016-sharpmask} and InstanceFCN \cite{dai2016instance}. We also compare to the hand-crafted system MCG \cite{pont2017multiscale}. Despite a comparison to  bounding box methods does not seem fair, as explained in Sec.~\ref{sec:rWObjectProposals}, we also compare to the very fast bounding box systems BING and EdgeBoxes. A comparison to instance segmentation methods is not fair as well due to their class-specific nature as explained in the introduction. As described in Sec.~\ref{sec:arch}, we will compare three versions of AttentionMask (AttentionMask$^{8}_{128}$, AttentionMask$^{8}_{192}$, and AttentionMask$^{16}_{192}$) that differ in the scales used. We did not include the results of FastMask with PVANet as base net, as the model is not publicly available.
\begin{figure}[tb]
\centering
\begin{tabular}{ccccc}
\subfloat{\includegraphics[width = .19\linewidth]{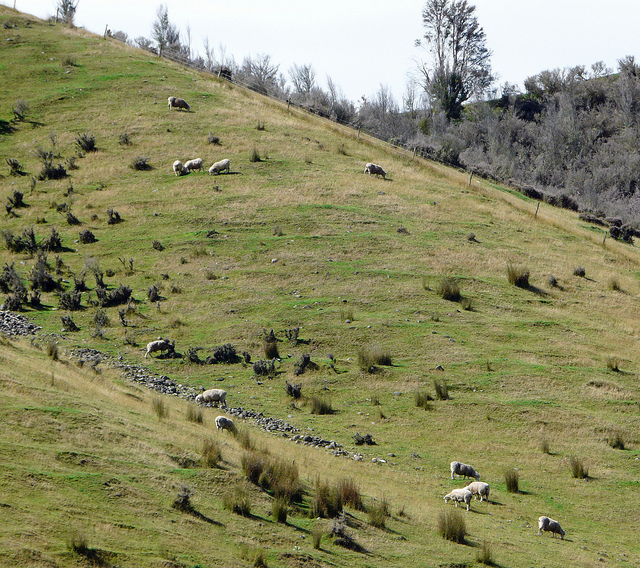}} &
\subfloat{\includegraphics[width = .19\linewidth]{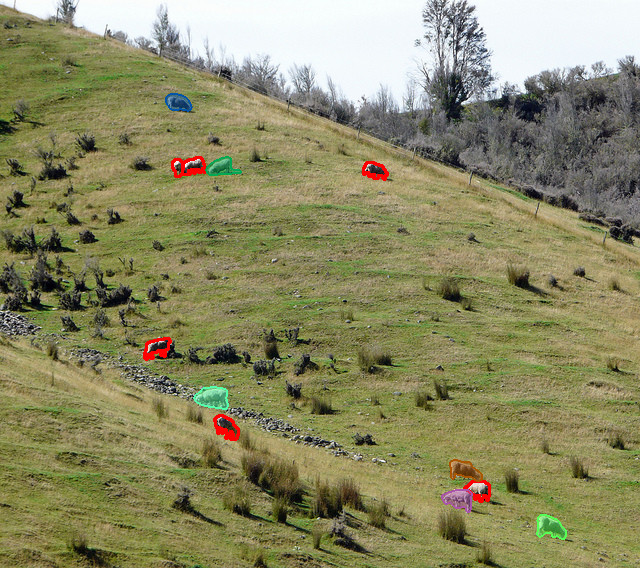}} &
\subfloat{\includegraphics[width = .19\linewidth]{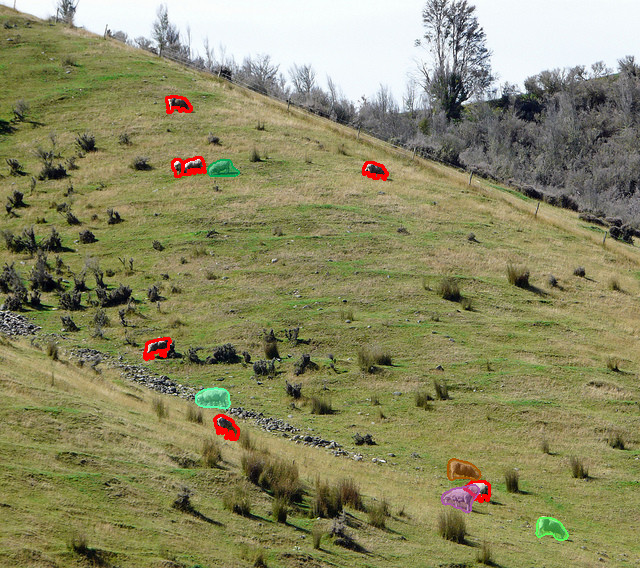}} &
\subfloat{\includegraphics[width = .19\linewidth]{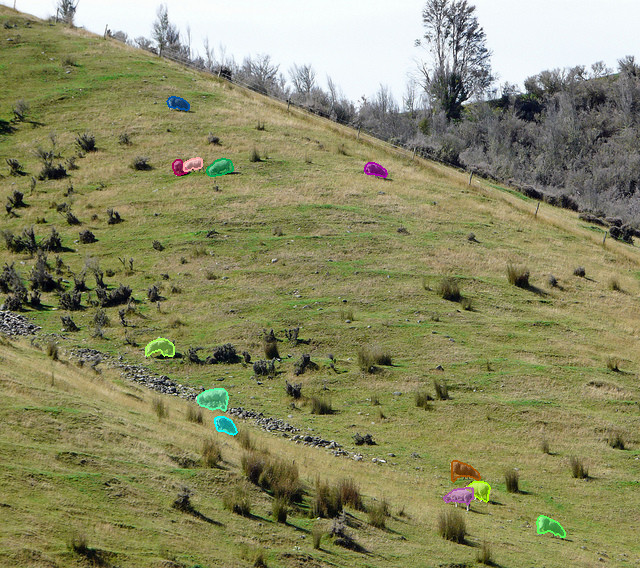}} &
\subfloat{\includegraphics[width = .19\linewidth]{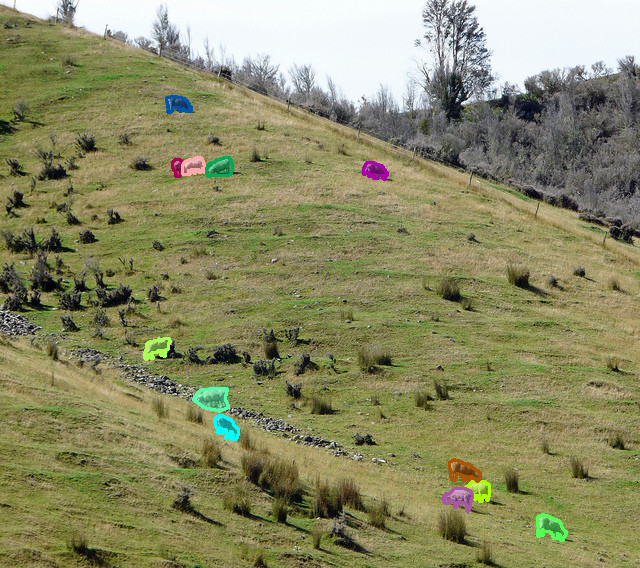}}\vspace{-3mm}\\
\subfloat{\includegraphics[width = .19\linewidth]{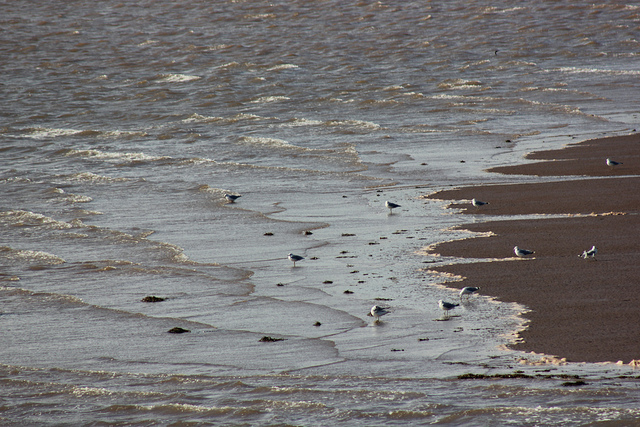}} &
\subfloat{\includegraphics[width = .19\linewidth]{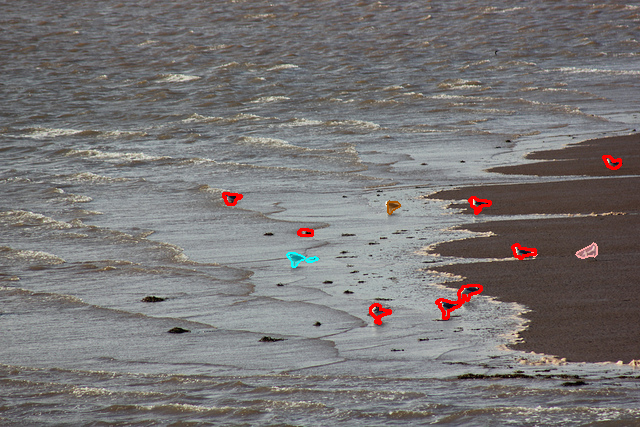}} &
\subfloat{\includegraphics[width = .19\linewidth]{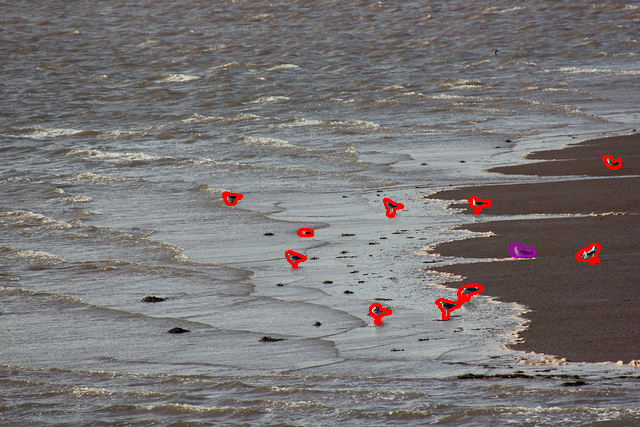}} &
\subfloat{\includegraphics[width = .19\linewidth]{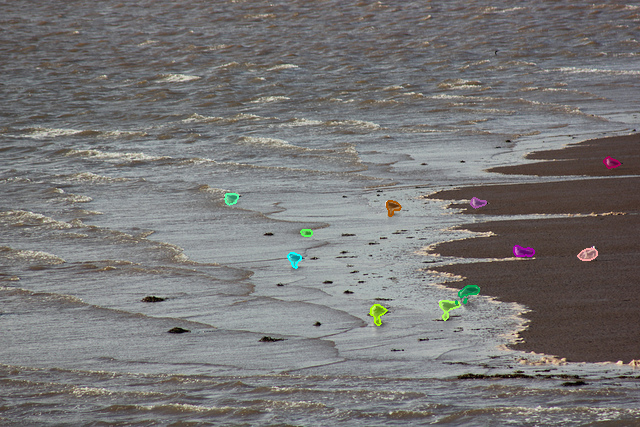}} &
\subfloat{\includegraphics[width = .19\linewidth]{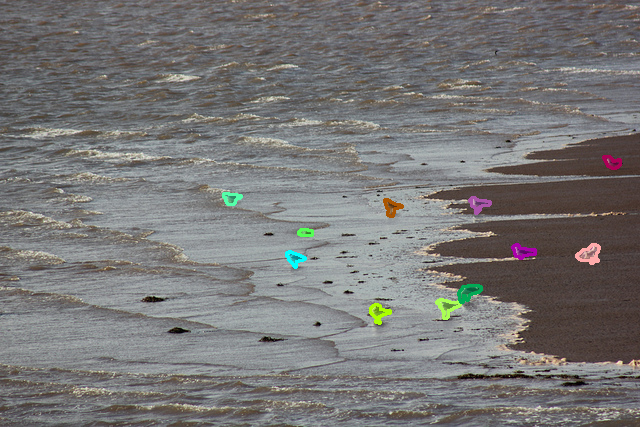}}\vspace{-3mm}\\
\subfloat{\includegraphics[width = .19\linewidth]{./img/COCO_val2014_000000011567}} &
\subfloat{\includegraphics[width = .19\linewidth]{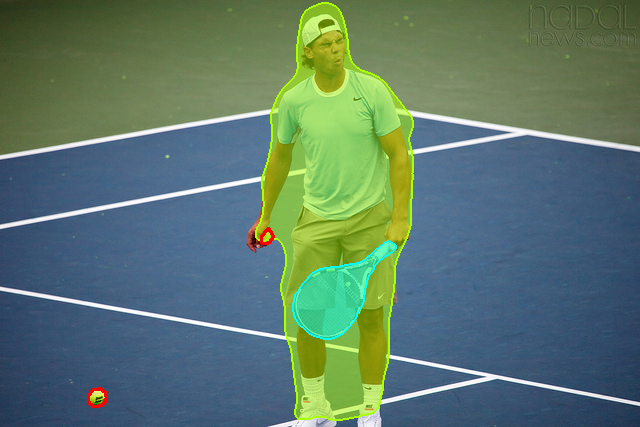}} &
\subfloat{\includegraphics[width = .19\linewidth]{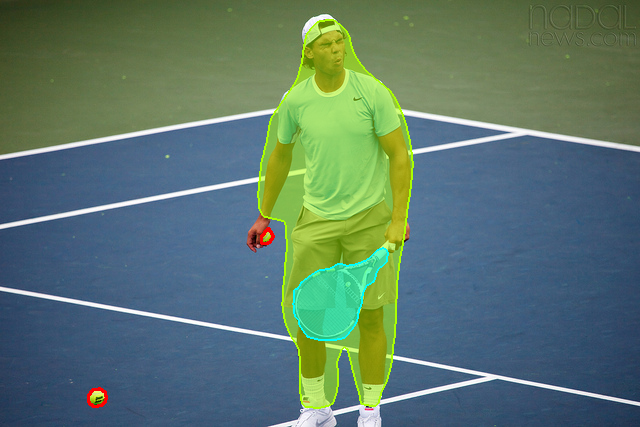}} &
\subfloat{\includegraphics[width = .19\linewidth]{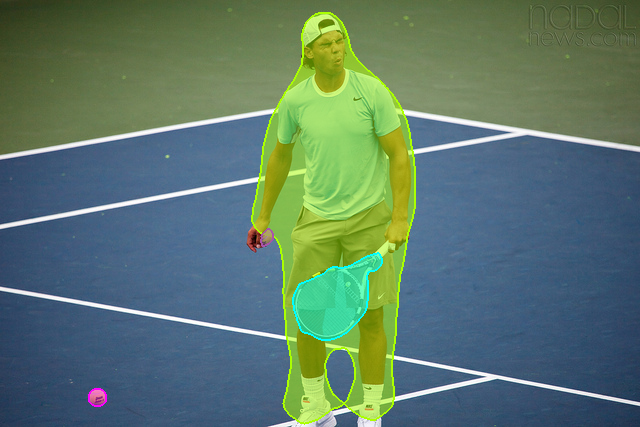}} &
\subfloat{\includegraphics[width = .19\linewidth]{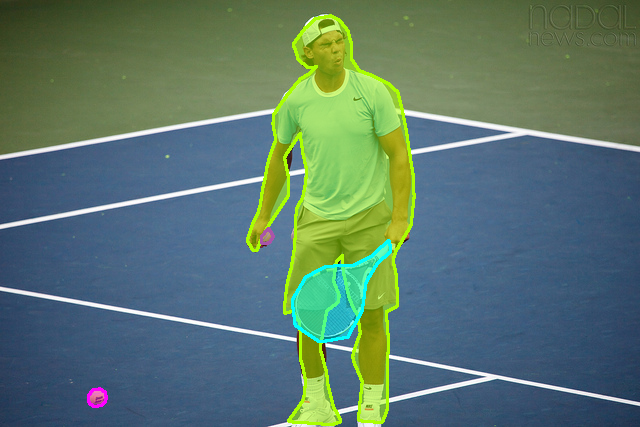}}\vspace{-3mm}\\
\subfloat{\includegraphics[width = .19\linewidth]{./img/COCO_val2014_000000047873}} &
\subfloat{\includegraphics[width = .19\linewidth]{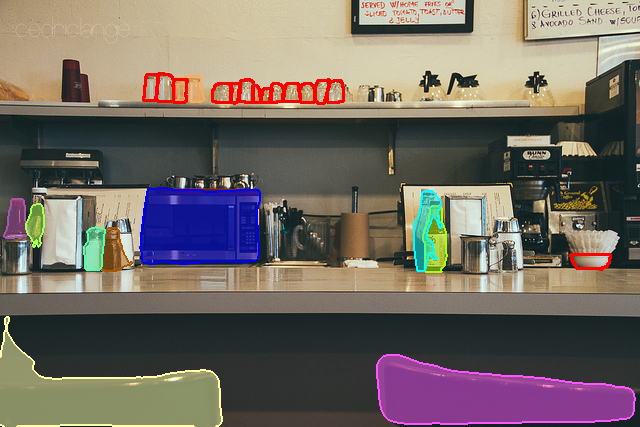}} &
\subfloat{\includegraphics[width = .19\linewidth]{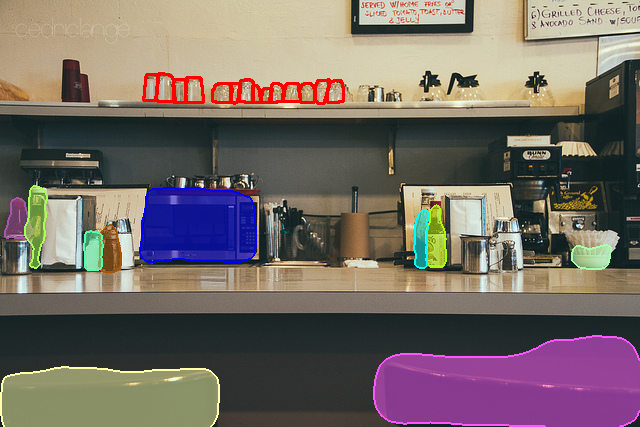}} &
\subfloat{\includegraphics[width = .19\linewidth]{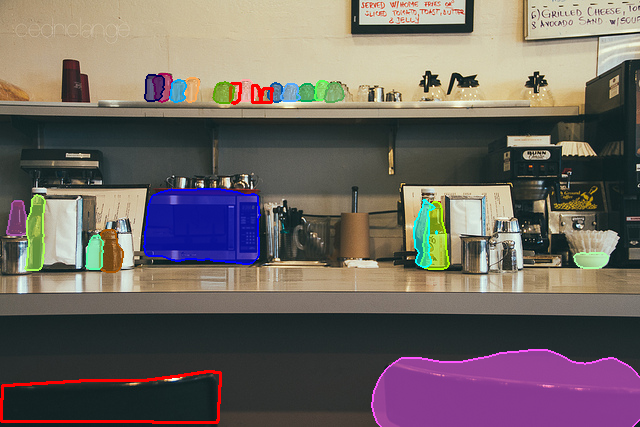}} &
\subfloat{\includegraphics[width = .19\linewidth]{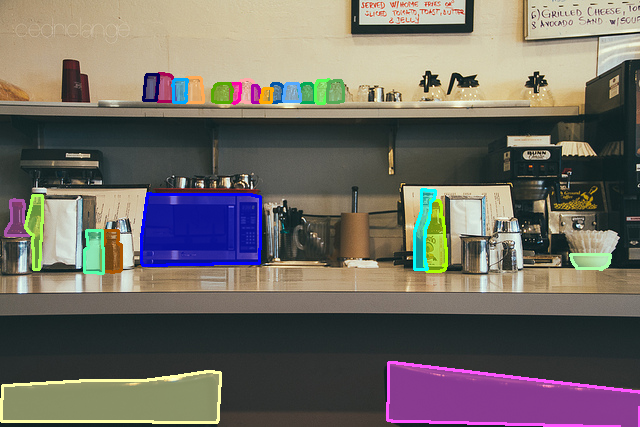}}\\
{\tiny Image}&{\tiny SharpMaskZoom~\cite{Pinheiro2016-sharpmask}} &{\tiny FastMask~\cite{Hu2017-fastmask}}&{\tiny AttentionMask$^{8}_{128}$}&{\tiny Ground Truth}
\end{tabular}
\caption{Qualitative results of SharpMaskZoom~\cite{Pinheiro2016-sharpmask}, FastMask~\cite{Hu2017-fastmask} and AttentionMask$^{8}_{128}$ on the MS COCO dataset. The filled colored contours denote found objects, while not filled red contours denote missed objects.
}
\label{fig:qualRes}
\end{figure}

\subsection{Qualitative Results}

The qualitative results visualized in Fig.~\ref{fig:qualRes} show the superior ability of AttentionMask to detect small objects compared to other state-of-the-art class-agnostic object proposal systems. For instance, the sheep (1st row) as well as the birds (2nd row) are all detected by AttentionMask. SharpMaskZoom and FastMask are not able to detect most of those small objects. Both images also show significant textures in some areas, which do not hamper the performance of the scale-specific objectness attention as the windows are still sampled at the correct locations. Returning to the earlier tennis example, the results in the third row show the ability of AttentionMask to detect both balls (bottom left and in his right hand). This emphasizes the effectiveness of the additional scale $\mathcal{S}_8$ and the attention focusing on the spots of the tennis balls in $\mathcal{S}_8$ as visualized in Fig.~\ref{fig:attExample}. Finally, the last row shows an example of a cluttered household environment with many small objects like the glasses on the shelf. AttentionMask is the only one of the three systems to detect almost all of them, although it is even hard for humans to detect the glasses in the input image given the low contrast. A more detailed version of this image highlighting the glasses can be seen in Fig.~\ref{fig:qualResCrop1}. More qualitative results can be found in the supplementary material.

\begin{table}[tbp]
\centering
\caption{
Results on MS COCO with pixel-precise proposals. S, M, L denote small, medium, large objects.  (Bounding box results in supplementary material)
}
\label{tab:resultsCOCO}
\begin{tabular}{@{}lccccccc@{}}
\toprule
Method &  AR@10 & AR@100  & AR@1k & AR$^S$@100  & AR$^M$@100  & AR$^L$@100 & Time \\ \midrule
MCG~\cite{pont2017multiscale} &  0.077 & 0.186  &  0.299 & -  &  - &-  &   $45s$                   \\
DeepMaskZoom~\cite{Pinheiro2016-sharpmask} & 0.151  &  0.286 &  0.371 &0.093  & 0.389  & 0.466  & $1.35s$                     \\
SharpMask~\cite{Pinheiro2016-sharpmask} &  0.154 & 0.278  &  0.360 &  0.035 & 0.399  & 0.513 &$1.03s$                      \\
SharpMaskZoom~\cite{Pinheiro2016-sharpmask} & 0.156  &  0.304 & 0.401  & 0.099  & 0.412  &  0.495 & $2.02s$                     \\ 
InstanceFCN~\cite{dai2016instance} & 0.166  &  0.317 & 0.392  & -  & -  &  - & $1.50s$                     \\ 
FastMask~\cite{Hu2017-fastmask} & 0.169  &  0.313 & 0.406  & 0.106  & 0.406  & 0.517 & $0.33s$                     \\ \midrule
AttentionMask$^{8}_{128}$ & 0.180  &  0.349 & 0.444  & \textbf{0.162}  & 0.421  & 0.560 & $0.22s$                     \\ 
AttentionMask$^{8}_{192}$ & \textbf{0.183}  &  \textbf{0.355} & \textbf{0.450}  & 0.157  & 0.426  & 0.590 & $0.22s$                     \\
AttentionMask$^{16}_{192}$ & 0.176  &  0.336 & 0.412  & 0.097  & \textbf{0.438}  & \textbf{0.594} & $\mathbf{0.21s}$ \\\bottomrule
\end{tabular}
\end{table}

\subsection{Quantitative Results}
\begin{figure}[tb]
    \centering
    \subfloat[AR$^S$]{\includegraphics[width = .3\linewidth]{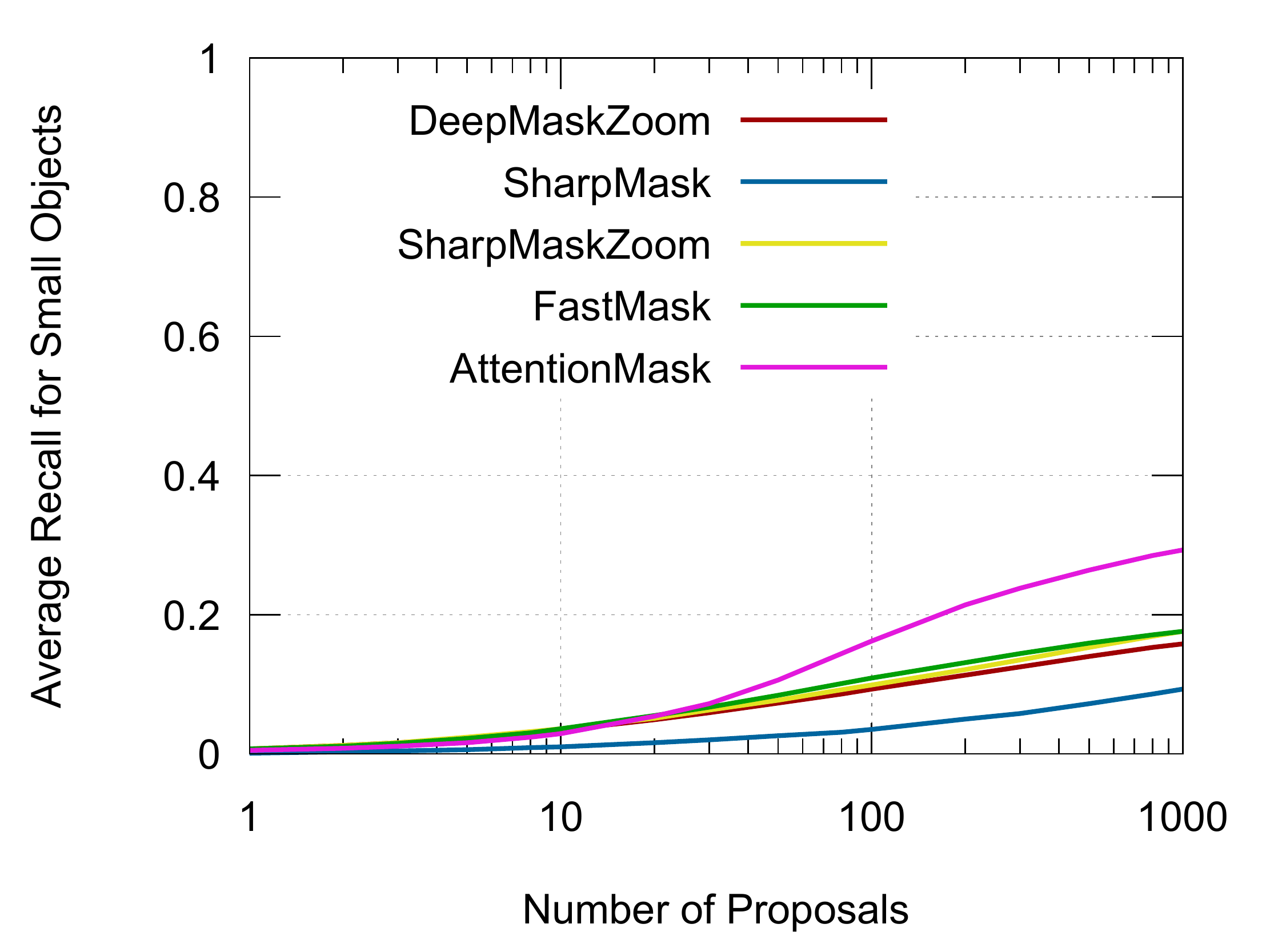}\label{fig:plotARS}}~
    \subfloat[Recall@100]{\includegraphics[width = .3\linewidth]{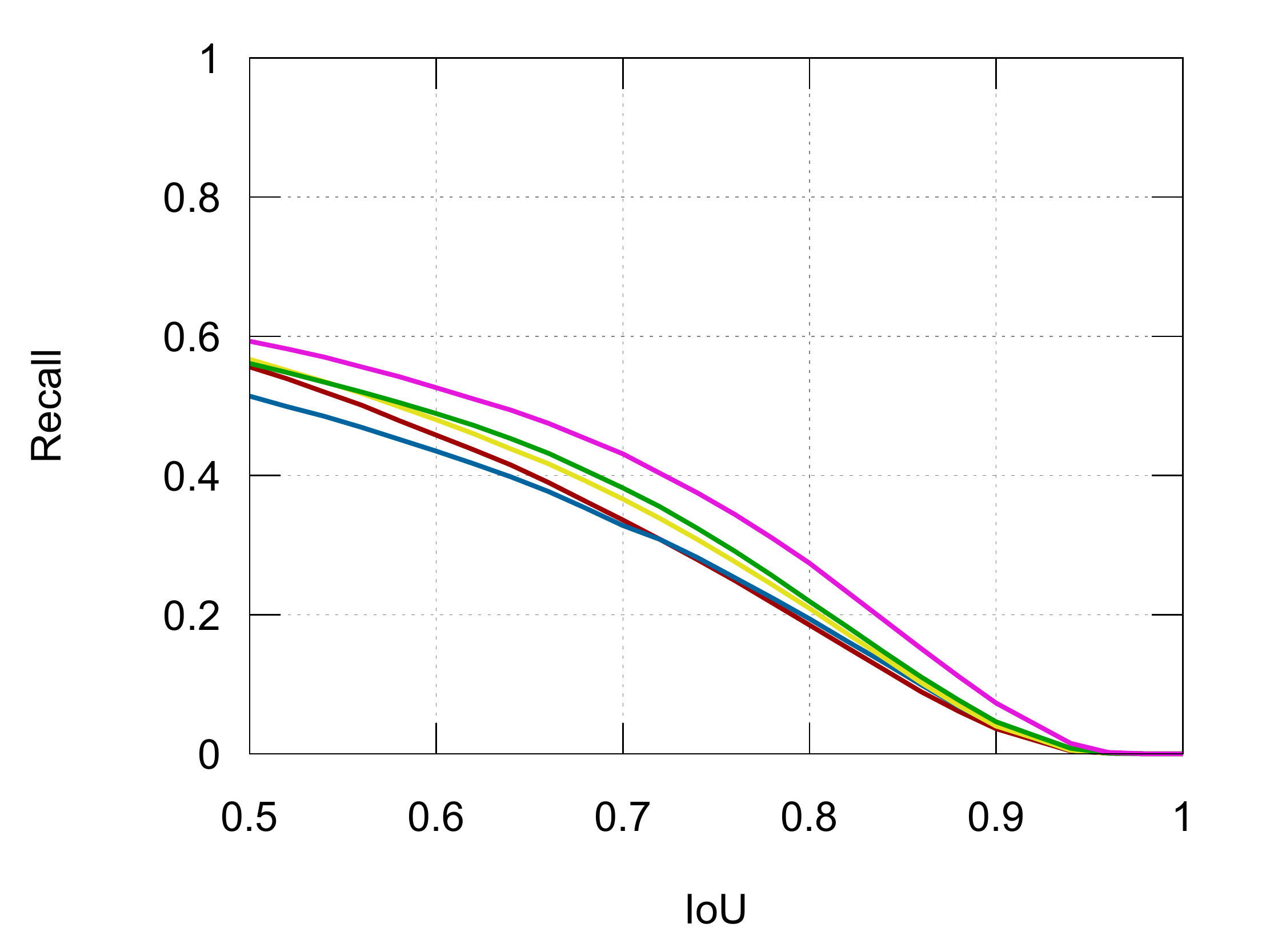}\label{fig:plotIoU}}~
    \subfloat[Inference Runtime]{\includegraphics[width = .3\linewidth]{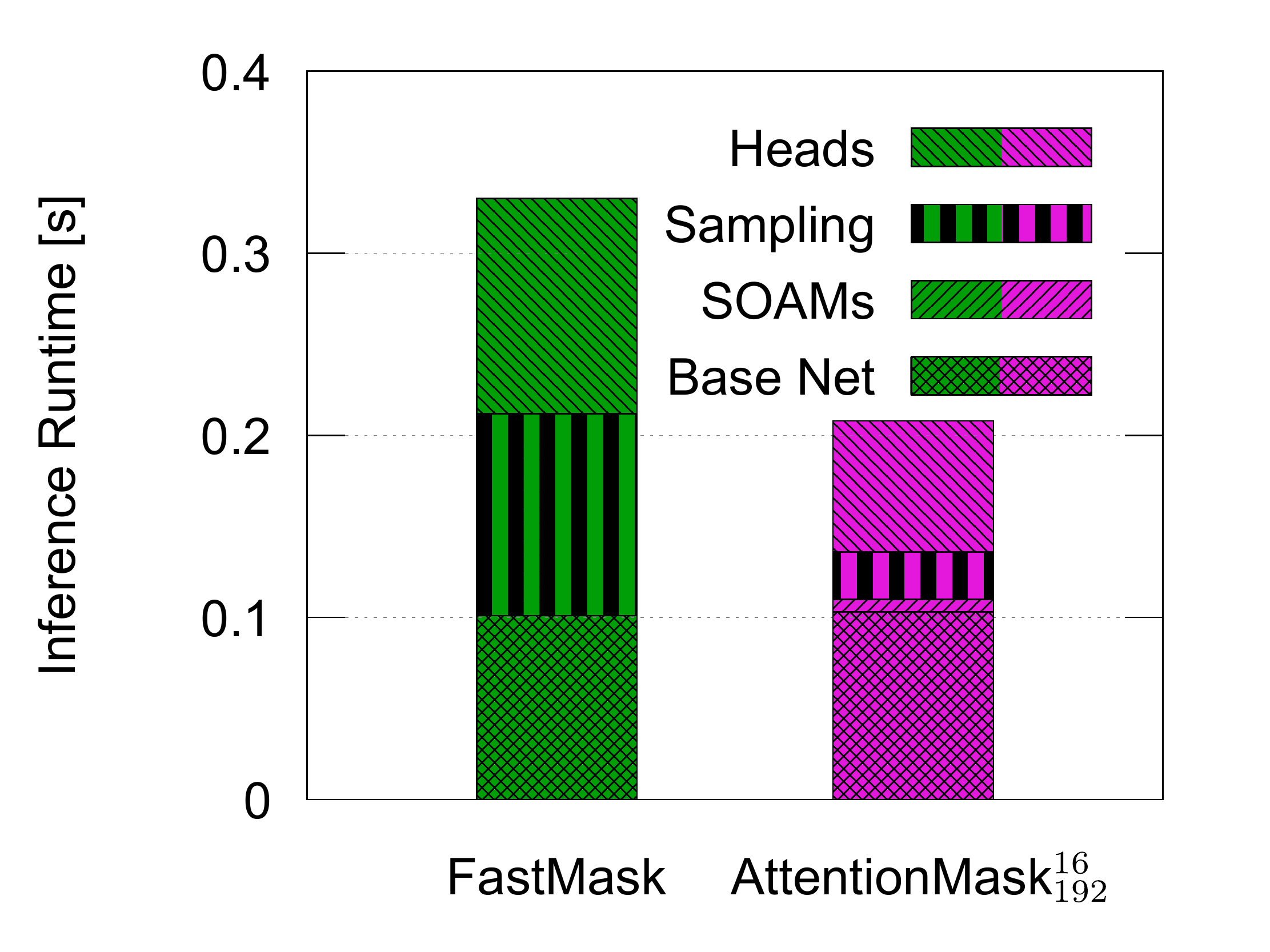}\label{fig:plotSpeed}}
            
    \caption{Detailed analysis of (a) AR$^S$ over different number of proposals, (b) recall at 100 proposals for different IoU values and (c) detailed runtime for inference.}\label{fig:plots}
\end{figure}

The findings of the qualitative results, which showed that AttentionMask is significantly better in detecting small objects, can be confirmed by the quantitative results. In addition, AttentionMask outperforms all state-of-the-art methods in all 7 categories, including inference runtime. Tab.~\ref{tab:resultsCOCO} shows the results of the different state-of-the-art methods and the three versions of AttentionMask. It is clearly visible that the two versions of AttentionMask that use $\mathcal{S}_8$ perform much better on small objects than any other method. The gain of AttentionMask$^{8}_{128}$, e.g., in AR$^S$@100 compared to FastMask is $53\%$ and compared to SharpMaskZoom even $59\%$ (all using 8 scales). A more detailed evaluation on small objects across different numbers of proposals is given in Tab.~\ref{tab:resultsSmallObj} and Fig.~\ref{fig:plotARS}. Both show that AttentionMask$^{8}_{128}$ significantly outperforms state-of-the-art methods on small objects across different numbers of proposals. As Fig.~\ref{fig:plotIoU} and Tab.~\ref{tab:resultsCOCO} show, across all scales the results of the three AttentionMask versions significantly outperform state-of-the-art as well. Compared to FastMask, SharpMaskZoom and InstanceFCN, the increase in AR@100 for AttentionMask$^{8}_{128}$, e.g., is between $10\%$ and $15\%$. Similar observation can be made for AR@10 and AR@1000.

\begin{table}[tb]
\parbox{.58\linewidth}{
\centering
\caption{Detailed results on small objects of the MS COCO dataset.}
\label{tab:resultsSmallObj}
\begin{tabular}{@{}lccc@{}}
\toprule
Method & AR$^S$@10 & AR$^S$@100 & AR$^S$@1k \\ \midrule
DeepMaskZoom~\cite{Pinheiro2015-deepmask,Pinheiro2016-sharpmask}&  0.034 &  0.093 &  0.158  \\
SharpMask~\cite{Pinheiro2016-sharpmask} &  0.010 &  0.035 &  0.093   \\
SharpMaskZoom~\cite{Pinheiro2016-sharpmask}  &  \textbf{0.036} &  0.099 &  0.176  \\
FastMask~\cite{Hu2017-fastmask}&  \textbf{0.036}   &  0.109  &  0.176 \\ 
AttentionMask$^{8}_{128}$&  0.029   &  \textbf{0.162}  &  \textbf{0.293} \\ \bottomrule
\end{tabular}
}
\hfill
\parbox{.35\linewidth}{
\centering
\caption{Analysis of the scale-specific objectness attention at different scales.}
\label{tab:evalAtt}
\begin{tabular}{@{}p{.45\linewidth}cc@{}}
\toprule
Scale & Recall & Pruned \\ \midrule
$\mathcal{S}_8$ & 0.781 & 0.939 \\ 
$\mathcal{S}_{16}$ & 0.831 & 0.924 \\
$\mathcal{S}_{32}$ & 0.982 & 0.640 \\
$\mathcal{S}_{64}$ & 0.969 & 0.594 \\
$\mathcal{S}_{128}$ & 0.918 & 0.704 \\ \bottomrule
\end{tabular}
}
\end{table}

In addition to outperforming state-of-the-art in terms of AR, the runtime for AttentionMask$^{8}_{128}$ is reduced compared to FastMask ($-33\%$), SharpMaskZoom ($-89\%$) and others without using a specialized base net as proposed in \cite{Hu2017-fastmask}. Compared to the efficient bounding box systems BING and EdgeBoxes (detailed numbers in supplementary material), our approach shows significantly better results (0.508 AR@100) compared to BING (0.084 AR@100) and EdgeBoxes (0.174 AR@100) on bounding boxes. Thus, despite BING and EdgeBoxes being fast as well ($0.20s$ for BING and $0.31s$ for EdgeBoxes), AttentionMask shows a more favorable trade-off between state-of-the-art performance and fast execution.

Adding $\mathcal{S}_{192}$ to AttentionMask does not improve the results significantly (AR@100 from 0.349 to 0.355). As expected, mostly large sized objects are added due to the addition of $\mathcal{S}_{192}$ (AR$^L$@100 from 0.560 to 0.590). The runtime for inference stays about the same compared to AttentionMask$^{8}_{128}$.  Thus, adding a smaller scale like $\mathcal{S}_{8}$ as introduced here is significantly more favorable.

\subsection{Ablation Experiments}
To analyze AttentionMask in more detail, we run multiple ablation experiments. For comparison between different structures of the SOAMs, we already provided results in Sec. \ref{sec:scaleSpecAtt} and Tab. \ref{tab:ablationAtt}, showing that more complex structures do not improve the results significantly. Similarly, we provided in Sec. \ref{sec:smallObjects} and Tab. \ref{tab:ablationS8} an analysis of different ways of integrating $\mathcal{S}_8$ to the system. Ablative experiments regarding different training strategies and the use of ground truth attention during training can be found in the supplementary material.

\textbf{Influence of $\mathcal{S}_8$.}
The results for AttentionMask$^{16}_{192}$ in Tab. \ref{tab:resultsCOCO} show the influence of the scale-specific objectness attention without adding $\mathcal{S}_8$.  Compared to FastMask the results are better across all scales (0.336 vs.0.313 for AR@100), emphasizing the positive effect of the introduced scale-specific objectness attention. Compared to AttentionMask$^{8}_{128}$, it is clearly visible that the numbers for small objects dropped as well as the overall result (0.336 vs.0.349 for AR@100). Thus, the addition of $\mathcal{S}_8$, only possible due to the time and memory efficiency gained by the introduced SOAMs, is crucial for the success of the system. Running FastMask with $\mathcal{S}_8$ is not possible due to the memory consumption.

\textbf{Runtime Evaluation.}
Evaluating the results of the scale-specific objectness attention shows that in AttentionMask$^{16}_{192}$, $86.7\%$ of the needed windows are sampled while reducing the overall amount of windows by $85.0\%$. This reduces the runtime for inference by $33\%$ compared to FastMask. Fig. \ref{fig:plotSpeed} gives a more detailed overview of the time consumed by the different stages: base net, SOAMs, window sampling and heads (segmentation + objectness). It is clearly visible that the time required for computing the attention maps is minimal ($0.007s$). However, the time required for window sampling and the heads is significantly reduced ($-0.131s$). This demonstrates the time efficiency due to the introduced scale-specific objectness attention and the new sampling scheme.

\textbf{Influence of scale-specific objectness attention.}
Further analyzing the scale-specific objectness attention at different scales (cf.~Tab.~\ref{tab:evalAtt}) shows the intended effect of pruning many unnecessary windows early in the process. For instance, at $\mathcal{S}_{16}$, $92.4\%$ of the windows are pruned, leading to a reduction of 1791 windows for an image of size $800 \times 600$. However, the recall, necessary windows that are found, is still at $83.1\%$. Thus, only few objects are lost through the scale-specific objectness attention. A similar behavior can be observed at other scales. Across $\mathcal{S}_8$ to $\mathcal{S}_{128}$  $91.7\%$ of the windows are pruned, leading to 9420 pruned windows in the example above. Still, the recall is at $80.3\%$.

\section{Conclusions}
\label{sec:conclusion}
State-of-the-art class-agnostic object proposal systems are usually based on an inefficient window sampling. In this paper, we have introduced scale-specific objectness attention to distinguish very early between promising windows containing an object and background windows. We have shown that this reduces the number of analyzed windows drastically ($-91.7\%$), freeing up resources for other system parts. Here, we have exploited these resources for addressing a second problem of current class-agnostic object proposal systems: the meager performance for small objects. This is especially important in real-world applications, since we are surrounded by many small objects. We added an additional module to our system, which significantly improves the detection of small objects, while across all scales outperforming state-of-the-art methods as well. Note that adding this module was only possible due to the resources freed by using scale-specific objectness attention. Our results clearly show the improvement in runtime ($33\%$ faster) and recall for small objects ($+53\%$) as well as across all scales ($+8\%$ to $+11\%$), which makes the approach especially suitable for extensions to video object proposal generation and the processing of real-world data.

\bibliographystyle{splncs04}
\bibliography{lit}
\end{document}